\documentclass{article}
\usepackage{spconf,amsmath,graphicx}
\usepackage{subfigure}
\usepackage{threeparttable}
\usepackage{cite}

\title{Pruning Large Language Models via Accuracy Predictor}

\name{Yupeng Ji$^{1}$\sthanks{Corresponding Author}, Yibo Cao$^{2}$, Jiucai Liu$^{3}$}
  
\address{$^{1,3}$ Tongji University, Shanghai, China \\ $^{2}$Chongqing University, Chongqing, China}

\begin{document}
\maketitle 


%
%

%

%
\begin{abstract}

Large language models(LLMs) containing tens of billions of parameters (or even more) have demonstrated impressive capabilities in various NLP tasks. However, substantial model size poses challenges to training, inference, and deployment so that it is necessary to compress the model. At present, most model compression for LLMs requires manual design of pruning features, which has problems such as complex optimization pipeline and difficulty in retaining the capabilities of certain parts of the model.Therefore, we propose a novel pruning approach: firstly, a training set of a certain number of architecture-accuracy pairs is established, and then a non-neural model is trained as an accuracy predictor. Using the accuracy predictor to further optimize the search space and search, the optimal model can be automatically selected. Experiments show that our proposed approach is effective and efficient. Compared with the baseline, the perplexity(PPL) on Wikitext2 and PTB dropped by 9.48\% and 5,76\% respectively, and the average accuracy of MMLU increased by 6.28\%.
\end{abstract}
\section{Introduction}
\label{sec:intro}

Recently, LLMs has shown impressive reasoning and generation capabilities in various NLP tasks, and these capabilities further enhance as the number of model parameters increases\cite{gpt3,chatgpt,llama1,llama2}. However, the huge computational and memory requirements are still a major obstacle to wider application, so it is necessary to compress LLMs to reduce costs~\cite{li2023auto,liu2023norm,li2022shadow,li2022SFF,li2022self,li2022tf}. Some works for LLMs compression currently focus on model quantization\cite{quant_1, quant_2,dong2023emq}, which is the process of quantizing model parameters into low-bit level representions. 

In face, another commonly used method for model compression is network pruning\cite{prune_survey}, which reduces the size of the model by deleting some unimportant weights. Although some pruning work for LLMs has made some progress, pruning strategies usually require manual design\cite{llm_pruner, wanda, sparsegpt}. Specifically, the pruning ratio of the global or each layer is preset, which causes the pruning results to depend on the set of hyperparameters. On the other hand, due to the model size and wide range of applications of LLMs, manual design of pruning strategies still faces challenges such as  complex pipelines and suitabilities for downstream tasks.

Neural architecture search (NAS)\cite{nas,linas2,li2021nas,dong2023diswot,dong2023rd,lichengp}, which aims to automatically find neural network architecture,  has been applied to model pruning due to its effectiveness and simplicity. One common approach is using an accuracy predictor to predict the accuracy of the model architecture to be pruned in the search space, which can save the cost of evaluating these candidate architectures\cite{gbdt_acc,wei2022convformer}. The key of this approach is training the accuracy predictor and the required dataset. Accuracy predictors among previous works are basically based on neural network, although effictive, it still requires a large number of architecture-accuracy pairs and careful design. However, for LLMs, a single evaluation also requires certain computing resources, so it is unrealistic to build tens of thousands of architecture-accuracy pairs.

In this paper, we propose an alternative approach that first builds a certain number of architecture-accuracy pairs for LLMs(usually a few hundred), and then trains an accuracy predictor based on non-nerural models(e.g., tree based models) to guide LLMs structured pruning. The specific algorithm is as follows:
(1) Combined with expert knowledge to limit the scope of some features(such as layer type, layer ID and so on), narrow the search space and perform random sampling to obtain the model architectures to be pruned.
(2) According to the pruning requirements of each model architectures,  prune the model, and evaluate the pruned model from multiple aspects to obtain the accuracy. Then train an accuracy predictor based on gradient boosting decision trees (GBDT)\cite{gbdt}.
(3) Use the trained GBDT model to predict more architectures in the search space, and architectures with top predicted accuracy are selected for further evaluation.

The main contributions of our approach are as follows:

(1) Explore the use of GBDT as an accuracy predictor, establish the relationship between the model architecture to be pruned and model performance, and improve the efficiency of searching model architecture.

(2) Use the trained accuracy predictor to further guide the process of model pruning, and conduct a refined search based on the required performance of a certain aspect of the model. It solves the problem of complex optimization pipeline caused by manually designing pruning features, making it easier to find the optimal model architecture.

\begin{figure}[h]
    \centering
    \includegraphics[width=0.5\textwidth]{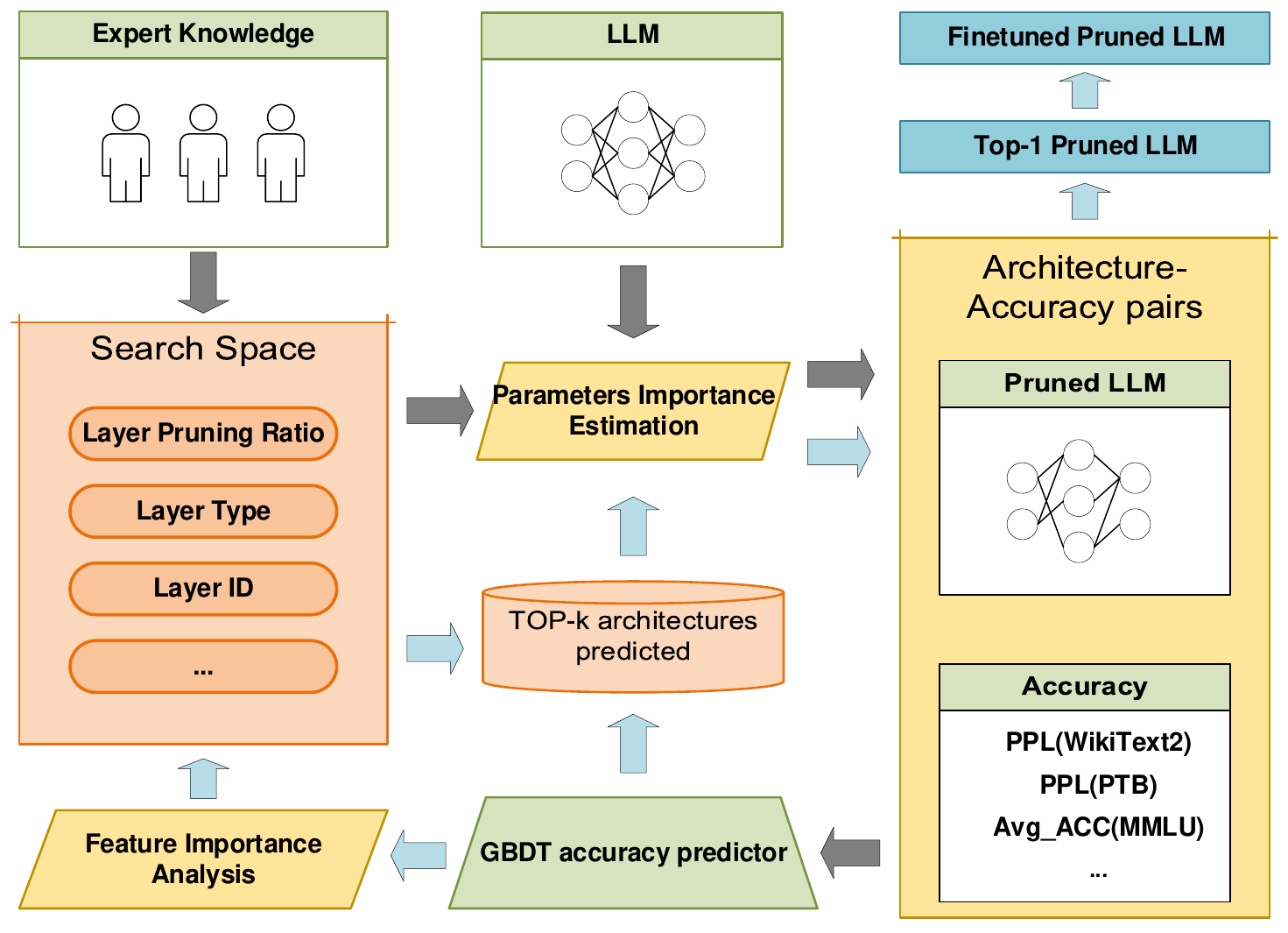}
    \caption{The overview of our approach}
    \label{fig:methodology}
\end{figure}

\section{methodology}
\label{sec:format}
Sec.\ref{build_pairs} describes the specific process of building architecture-accuracy pairs and accuracy predictors including how to narrow the search space of pruning, how to evaluate the parameters importance of the model, how to evaluate the pruned model, and how to train the accuracy predictor. Sec.\ref{gbdt_for_NAS_search} introduces the use of GBDT as a predictor for architecture search, which make the overall search process more efficient and effective. Fig.\ref{fig:methodology} shows the overview of our approach.

\subsection{Building Architecture-Accuracy Pairs} \label{build_pairs}

\textbf{Narrow the Search Space} Usually, we prune the model according to the predefined pruning ratio of each layer and evaluate the pruned model, so that we get a architecture-accuracy pair, but the problem is how to predefined the pruning ratio of each layer. It is unachievable to predict the accuracy for all the candidate architectures is costly for large search space in real applications.

Therefore, we have to limit the ID and type of LLM layers to be pruned, and set a range of the pruning ratio (the proportion of the layer with parameters equal to 0). Furthermore, adjacent $k$ layers can be considered to use the same pruning ratio. Finally, random sampling is performed to obtain the architectures.

\textbf{Parameters Importance Estimation} 


Consider a set of samples $\mathcal{D} = \{(x_i, y_i), i = 1,2,...,N\}$,  our goal is to remove the parameters that have the least impact, where the importance of the parameters can be represented by the deviation of the loss\cite{obd}. Specifically, as is shown in Eq.(\ref{eq.taylor_1}), we use Taylor Expansion to approximate the change in the loss $\Delta \mathcal{L}$ of pruning structures\cite{molchanov2019importance}.  

\begin{equation} \label{eq.taylor_1}
    \begin{aligned}
        \Delta \mathcal{L} 
        = \frac{\partial \mathcal{L}^T}{\partial w} \Delta w + \frac{1}{2} \Delta w^T \boldsymbol{H} \Delta w + R_2(w)
    \end{aligned}
\end{equation}

where $R_2(w)$ is remainders, $\frac{\partial \mathcal{L}^T}{\partial w}$ is the first-order gradient of loss function, and $\boldsymbol{H}$ is the Hessian matrix containing second-order derivatives. Although second-order Taylor expansion contains more information, it requires more additional memory and computational resources, which is more expensive for LLMs. On the contrary, the gradient information of first-order Taylor expansion can be obtained from backpropagation, which is more efficient. Therefore,  the first-order Taylor expansion is used for importance estimation.



\textbf{Evaluation of The Pruned LLMs} Currently, due to the recent emergence of LLMs and their excellent performance in multiple fields, scientific and comprehensive quantitative evaluation is still being explored\cite{explainability}. We evaluate LLMs from two aspects:

1) Generation ability: zero-shot perplexity(PPL) analysis on Wikitext2\cite{wikitext2} and PTB\cite{ptb}. The lower the value, the stronger the ability.

2) World knowledge and problem solving ability: five-shot accuracy analysis on Massive Multitask Language Understanding(MMLU)\cite{mmlu} which covers 57 subjects across STEM, the humanities, the social sciences, and more. The higher the value, the stronger the ability.

As LLMs evaluation mechanism continues to improve, more quantitative evaluation indicators can be added.

\textbf{Training Accuracy Predictor Based on GBDT} After obtaining $M$ architecture-accuracy pairs, as is shown in Table , we train an accuracy predictor based on GBDT with the goal of minimizing the difference between prediction accuracy and target accuracy as much as possible. It should be noted that some data preprocessing work is necessary, such as normalization.

So that, we can get architecture-accuracy pairs $\mathcal{C} = \{\boldsymbol{X_i}, \boldsymbol{Y_i}\}_{i=1}^{M}$, where $\boldsymbol{X_i} = (x_{i1}, x_{i2}, ..., x_{ip})$ is the architecture features, $p$ is its dimension, $\boldsymbol{Y_i} = (y_{i1}, y_{i2}, ..., y_{iq})$ is the accuracy of the pruned LLMs, $q$ is its dimension and $M$ is the number of architecture-accuracy pairs.

\subsection{Using GBDT for Search Space Search}
\label{gbdt_for_NAS_search}
One important reason for using GBDT as accuracy predictor is its interpretability. We can optimize search space according to the feature importance of the GBDT model.

Specifically, in order to maintain consistency with training architecture-accuracy pairs $\mathcal{C}$, we still need to comply with the requirements of Sec.\ref{build_pairs}. Then for features with higher importance, we try to limit the pruning ratio to a small range, while for features with lower importance, we can appropriately expand the range of pruning ratio to control the global pruning ratio of LLMs.

After determining the search space, we predict the accuracy of $L$ randomly sampled architectures. Then we evaluate these architectures with the top $K$ predicted accuracies and choose the best one.

\section{Experiments}
\label{experiment}

We first introduce the experiment setup in Sec.\ref{Experiment_Setup}. Then we carry out experiments in Sec.\ref{Experiment_results}, and verify the contribution of our approach. Finally, we try to restore the performance of pruned model in Sec.\ref{Experiments_qlora}.

\subsection{Experiment Setup}
\label{Experiment_Setup}
Our experiments are implemented with PyTorch  and the open source LLM structure used for testing is LLAMA2-7B\cite{llama2}. We select 10 randomly samples from Bookcorpus\cite{bookcorpus} for calculating the gradient. 

\subsection{Experiment Results and Analysis}
\label{Experiment_results}

\textbf{Build Architectures-Accuracy Pairs} There are a total of 32 LlamaDecoderLayers in LLAMA2-7B, we assume that only the LlamaAttention and LlamaMLP are pruned, the layer ID to be prued is \textit{[3,29]}\footnote{\textit{[3, 29]} represents a closed interval, that is, the value range is from 3 to 29}, the pruning ratio range is from 0.1 to 0.5, step is 0.1. In order to make the model more robust, during random sampling, we partially restrict the adjacent 4 layers to use the same pruning ratio, and some do not. The former sampled 400, and the latter sampled 125, for a total of 525 architecture-accuracy pairs, as is shown in Tab.\ref{table_architecture_accuracy_pairs}.

\begin{table}[t] 
\centering
\caption{Examples of architecture-accuracy pairs}
\label{table_architecture_accuracy_pairs}  
\resizebox{\linewidth}{!}{
\begin{tabular}{cccccccc}
    \hline\noalign{\smallskip}	
    $\boldsymbol x_{1}$ & $\boldsymbol x_{2}$ & $\boldsymbol x_{3}$ & ... & $\boldsymbol x_{54}$ & $\boldsymbol y_{1} $& $\boldsymbol y_{2} $&$ \boldsymbol y_{3} $  \\
    (Attention layer 3) & (Attention layer 4) & (Attention layer 5) & ... & $(MLP layer 29)$ & $(PPL_{Wikitext2})$ & $(PPL_{PTB})$ & $AVG\_ACC_{MMLU}$  \\
    \noalign{\smallskip}\hline\noalign{\smallskip}
    0.2 & 0.2 & 0.2 & ... & 0.5 & 45.3428 & 158.7869 & 0.1523  \\
    0.3  & 0.4 & 0.1 & ... & 0.2 & 83.4239 & 234.2987 & 0.0239  \\
    ... & ... & ... & ... & ... & ... & ... & ... \\
\noalign{\smallskip}\hline
\end{tabular}}
\end{table}

\textbf{Train GBDT and Predict} After building architectures-accuracy pairs, we divide to the training sets and the validation sets according to the ratio of 7:3, and mulit-output GBDT are trained. The metrics of validation sets are shown as Tab.\ref{table_gbdt}. 
\begin{table}[h] 
	\centering
	\caption{The metrics of validation sets}
	\label{table_gbdt}  
	\begin{tabular}{cc}
		\hline\noalign{\smallskip}	
		Algorithms & Values   \\
		\noalign{\smallskip}\hline\noalign{\smallskip}
		Mean Squared Error (MSE) &  0.1743  \\
		Root Mean Squared Error (RMSE) & 0.3861 \\
            Mean Absolute Error (MAE) & 0.2088 \\
            R-Square (R2) & 0.8040 \\
		\noalign{\smallskip}\hline
	\end{tabular}
\end{table} 

\begin{figure}[h]
    \centering
    \includegraphics[width=0.5\textwidth]{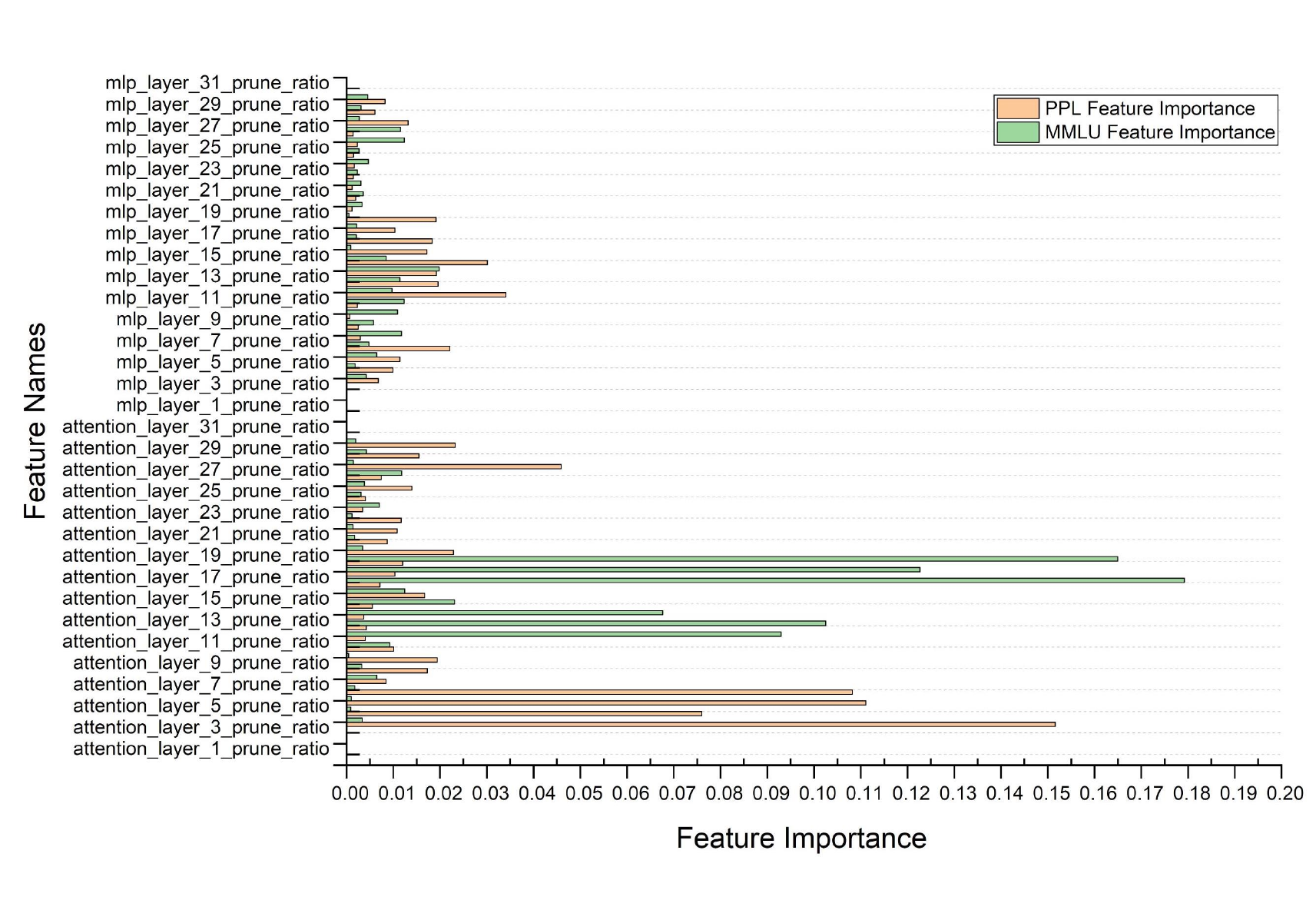}
    \caption{The feature importance for PPL and the average accuracy of MMLU}
    \label{fig:feature_importance}
\end{figure}

Then we calculate feature importance of PPL(Wikitext2 and PTB) and MMLU average accuracy, as is shown in Fig.\ref{fig:feature_importance}. It can be seen that the importance of the Attention layer is much higher than that of MLP. For PPL, the layers located at the head and tail are more important, while for MMLU, the importance of the middle layer is higher. This also reflects the different abilities of each layer in LLM. In addition, whether it is attention or mlp, the importance of the \textit{[19, 25]} layer is low. It may be that they represent some capabilities of the model that we have not assessed.

When predicting, we partially align with the training set, such as pruning only Attention and MLP layer, and keeping the first 3 layers and the last two layers unchanged. The difference is that we adjust the pruning ratio range based on the calculation results of feature importance. Specifically, the pruning ratio range of some Attention layers (\textit{[3,6]}, \textit{[11,18]}, \textit{[26,29]}) is narrowed to \textit{[0.1, 0.3]}, and the pruning ratio range of some MLP and Attention layers (\textit{[19, 25]}) is expanded to \textit{[0.2, 0.7]}.

\begin{figure*}[htbp]
  \centering
  \subfigure[Wikitext2]{\includegraphics[width=0.33\linewidth]{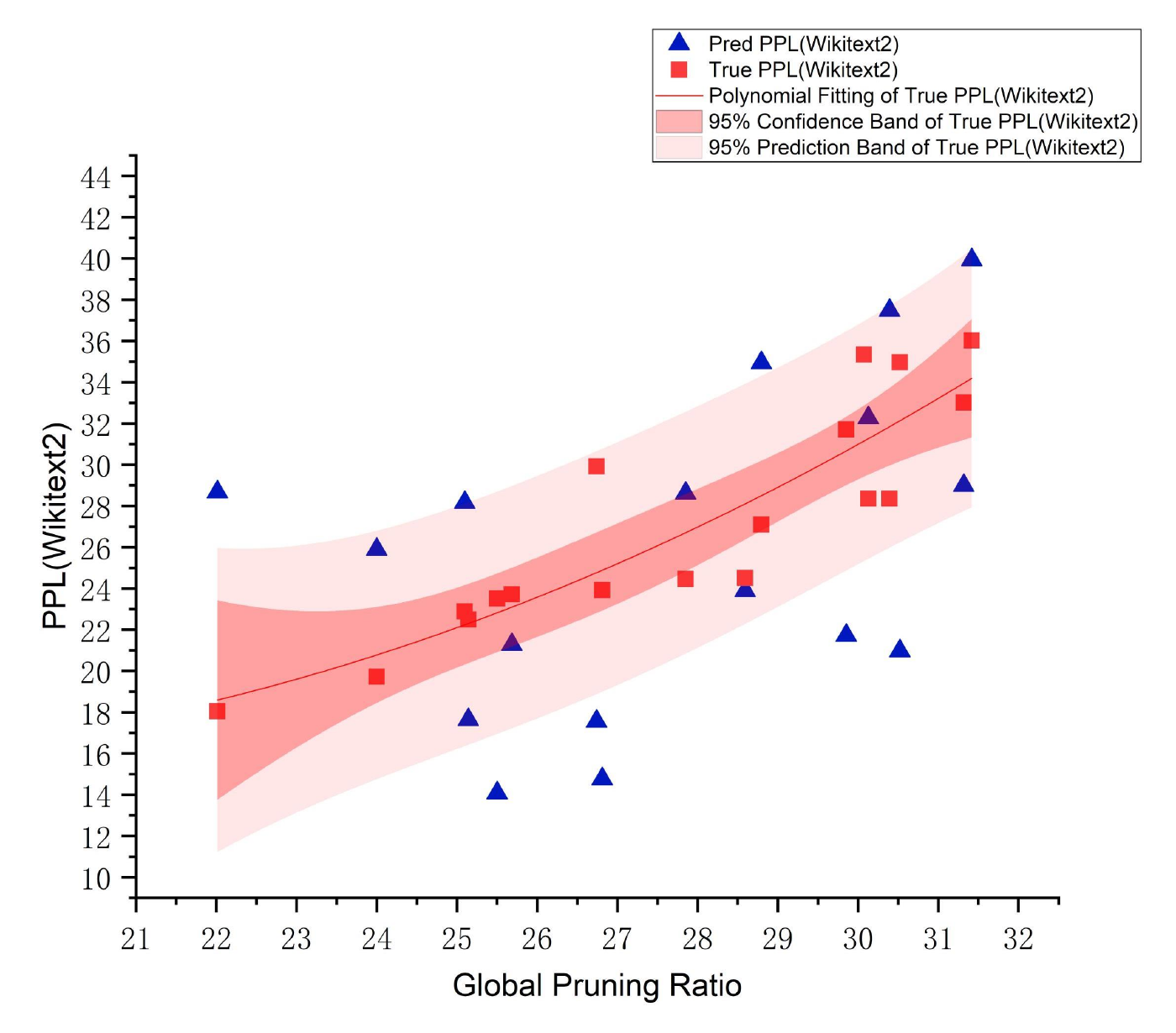}}
  \hfill
  \subfigure[PTB]{\includegraphics[width=0.33\linewidth]{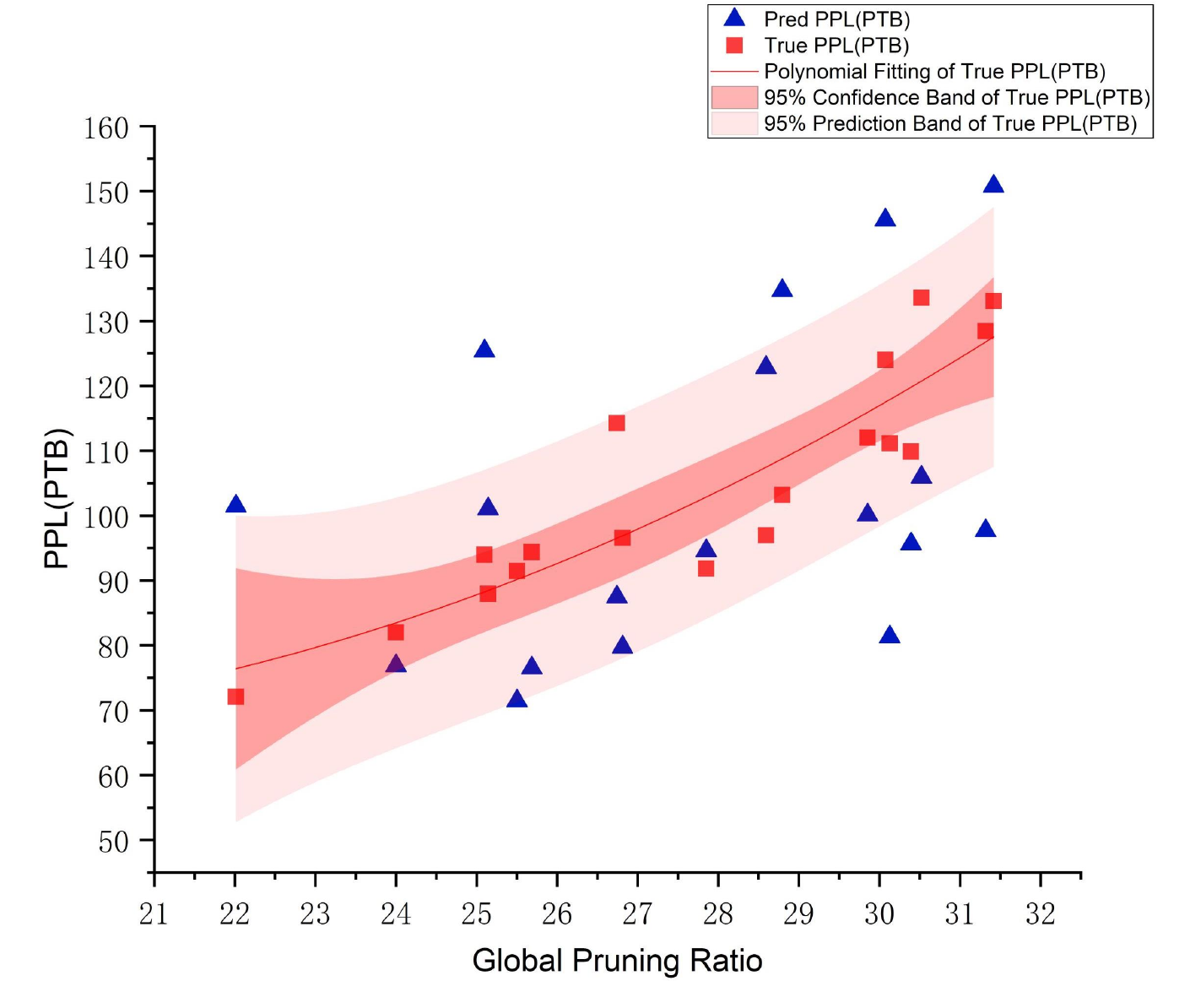}}
  \hfill
  \subfigure[MMLU]{\includegraphics[width=0.33\linewidth]{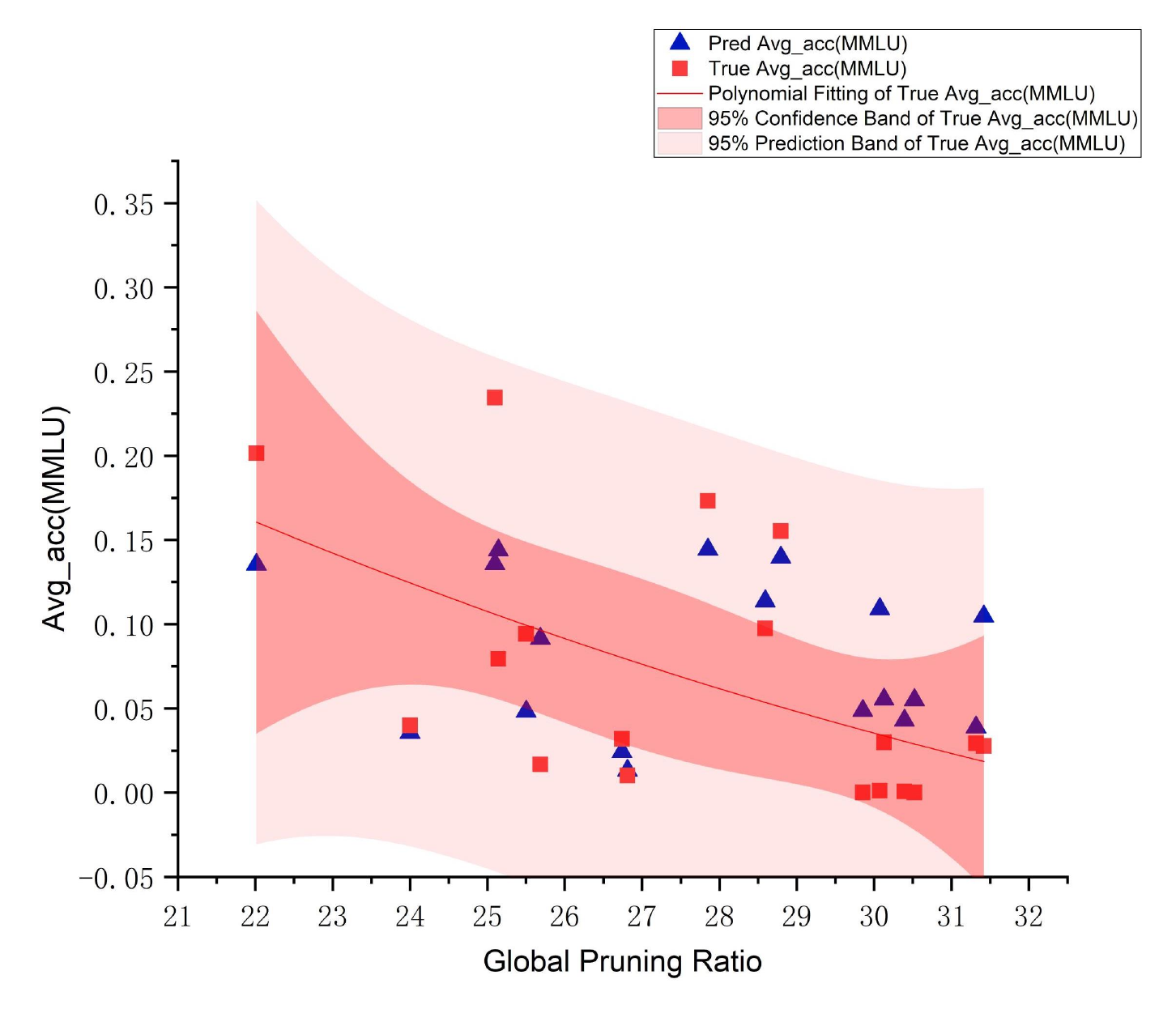}}
  \caption{Comparison of prediction accuracy using GBDT and true accuracy}
  \label{fig:gbdt_prediction}
\end{figure*}

\begin{figure*}[ht]
  \centering
  \subfigure[Wikitext2]{\includegraphics[width=0.33\linewidth]{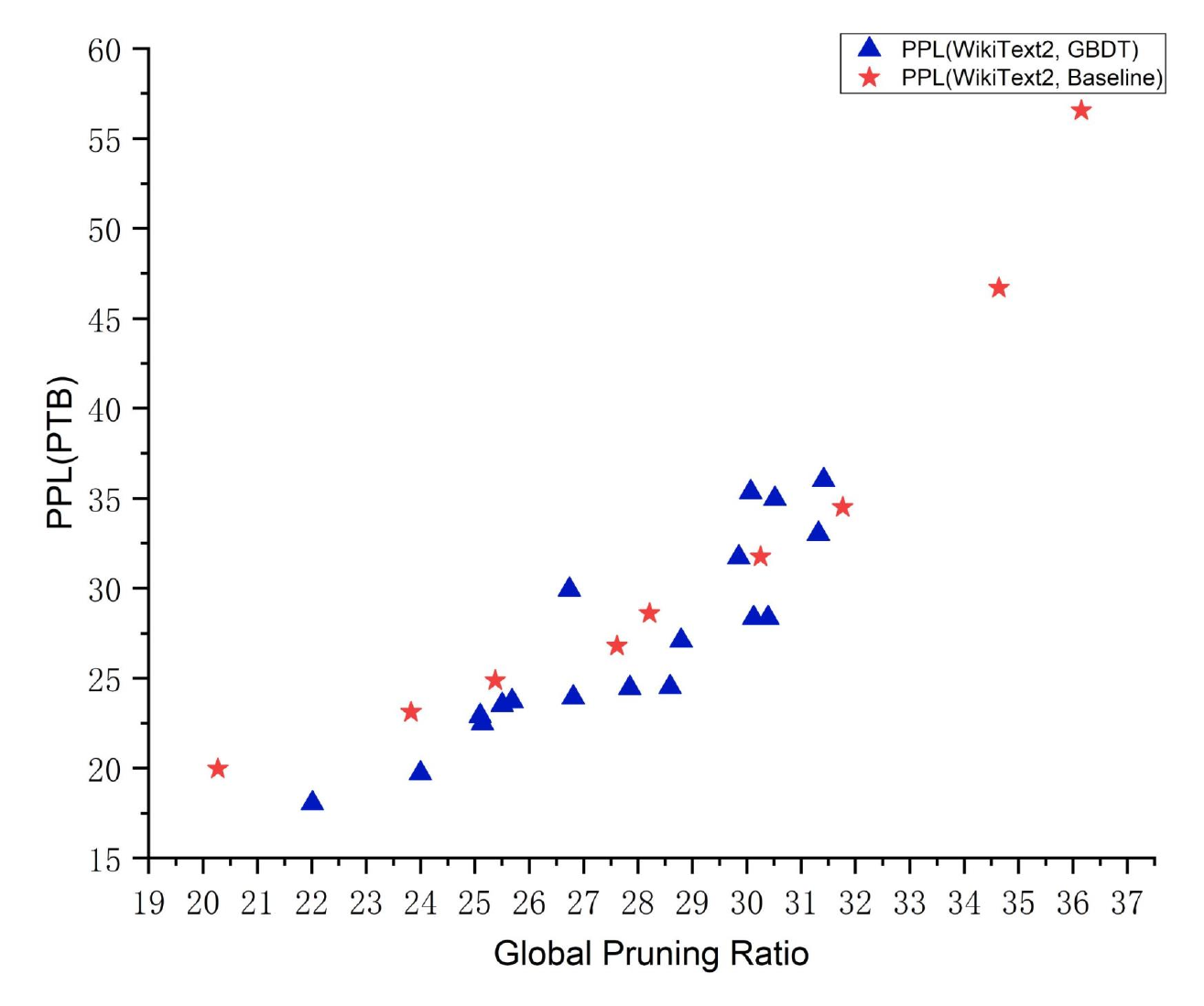}}
  \hfill
  \subfigure[PTB]{\includegraphics[width=0.33\linewidth]{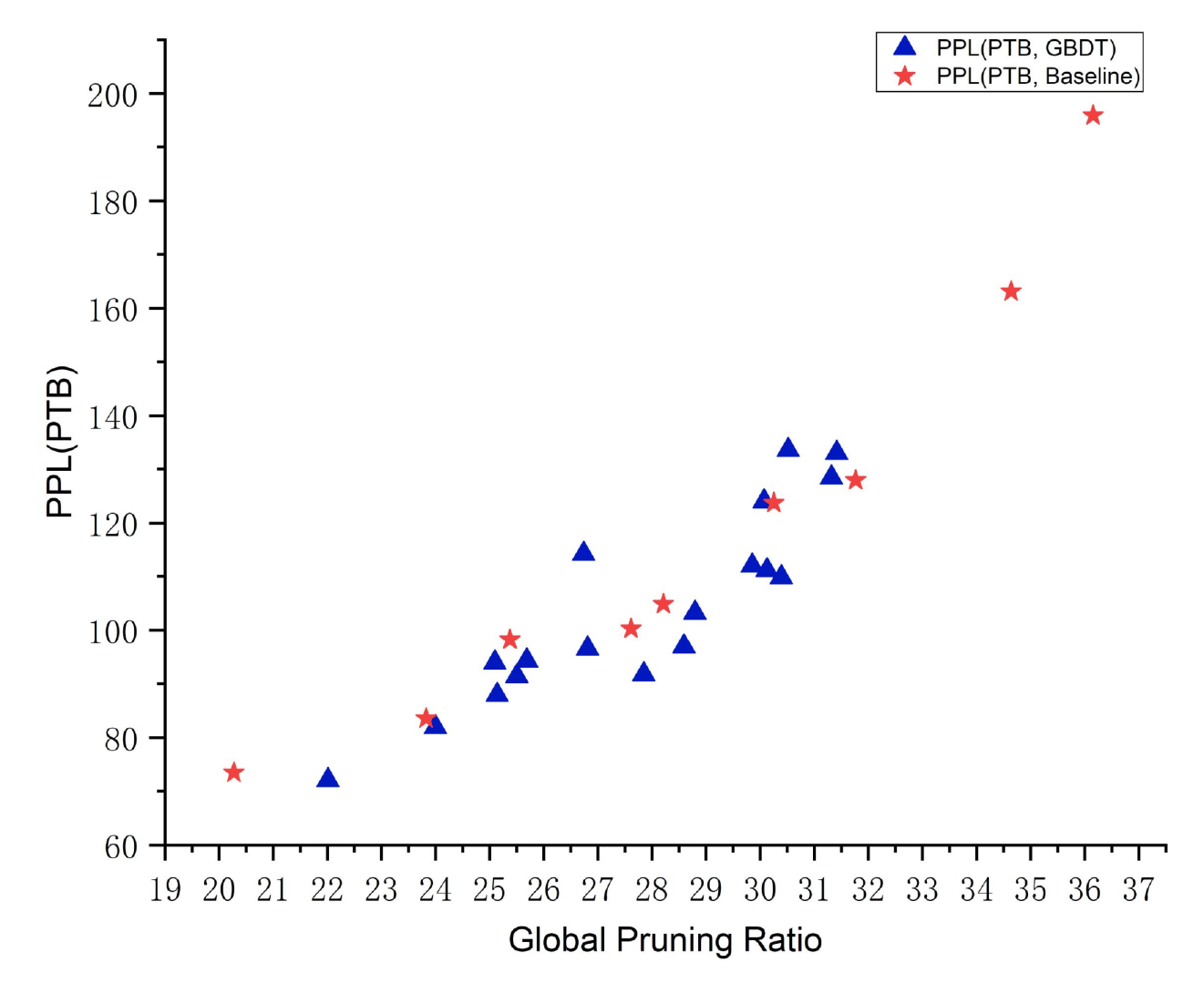}}
  \hfill
  \subfigure[MMLU]{\includegraphics[width=0.33\linewidth]{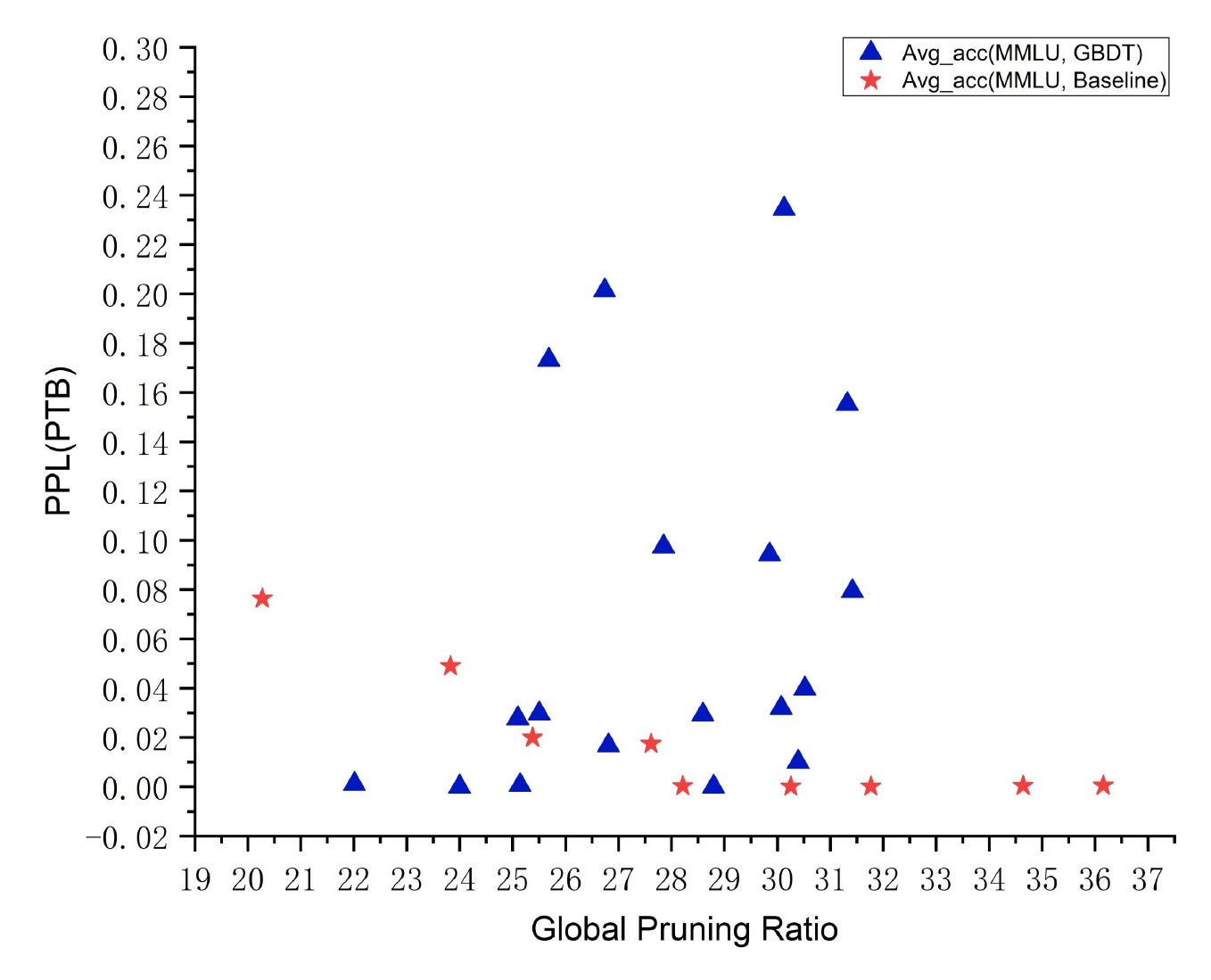}}
  \caption{Comparison Results of the pruned model choosed by GBDT and baseline}
  \label{fig:gbdt_baseline}
\end{figure*}

Then conducte random sampling, obtaine 10,000 model architectures, and make predictions. Based on different global pruning ratios, we selected 18 architectures with the best prediction results for evaluation. As shown in the Fig.\ref{fig:gbdt_prediction} , we performed a second-order polynomial fit on the true evaluation results of the architectures and established a 95\% confidence band and prediction band. It can be seen that the trained GBDT can be used as a predictor to predict unseen architectures, improving the efficiency of searching for better architectures in the search space.

We set the method based on LLM-pruner\cite{llm_pruner} as baseline. The only difference from our method is that the pruning ratio of each layer is set to the same. Filter according to the global pruning ratio, and the final result is shown in the Fig.\ref{fig:gbdt_baseline}. It is normal for the results to differ under almost the same global pruning ratio, as the specific pruning ratio of each layer vary. It can be seen that at almost the same global pruning ratio, the PPL value evaluated by  the best pruned model used GBDT on Wikitext2 is about 9.48\% lower than the baseline , on PTB is about 5.76\%. And the average accuracy evaluated on MMLU is about 6.28\% higher than the baseline. 

\subsection{Fast Recovery with QLoRA}
\label{Experiments_qlora}

Furthermore, we explore to restore model performance under limited data by Quantized Low-rank Approximation(QLoRA)\cite{qlora}. We selecte several pruned LLMs and collect 52k instruction samples. Tuning these samples requires merely 4 hours on a single GPU with only 2 epochs. As is shown in Tab.\ref{table_qlora}, the PPL(Wikitext2), PPL(PTB)  and Avg-acc(MMLU) increased by an average of 31.23\%, 37.5\% and 11.18\% respectively after the fine-tuning.

\begin{table}[h] 
	\centering
	\caption{Model performance before and after fine-tuning}
	\label{table_qlora}  
        \resizebox{\linewidth}{!}{
	\begin{tabular}{cccc}
		\hline\noalign{\smallskip}	
		Global Pruning Ratio & $PPL_{Wikiext2}$ & $PPL_{PTB}$  & $AVG\_ACC_{MMLU}$ \\
		\noalign{\smallskip}\hline\noalign{\smallskip}
		25.50\% & 23.50(\textbf{16.80})&91.43(\textbf{58.43})&0.0942(\textbf{0.1917}) \\
		25.69\% & 23.71(\textbf{18.71})&94.33(\textbf{61.96})&0.0168(\textbf{0.1982}) \\
            26.74\% & 29.92(\textbf{16.88})&114.24(\textbf{60.48})&0.0320(\textbf{0.1550}) \\
            26.81\% & 23.92(\textbf{16.16})&96.57(\textbf{58.76})&0.0101(\textbf{0.2043}) \\
            28.59\% & 24.51(\textbf{19.55})&96.95(\textbf{63.54})&0.0975(\textbf{0.2219}) \\
            28.80\% & 27.09(\textbf{17.20})&103.20(\textbf{62.09})&0.1552(\textbf{0.1214}) \\
            30.07\% & 35.34(\textbf{17.02})&124.00(\textbf{58.75})&0.0011(\textbf{0.1307}) \\
            30.52\% & 34.97(\textbf{30.98})&133.56(\textbf{110.01})&0.0000(\textbf{0.0109}) \\
		\noalign{\smallskip}\hline
	\end{tabular}
 }
 \begin{tablenotes}
			\footnotesize
			\item \textbf{Bold} represents the result after fine-tuning
\end{tablenotes}
\end{table}

\section{Conclusion}
\label{sec:conclusion}

In this paper, we propose a novel pruning approach using non-neural model for LLMs. We introduce GBDT into the architecture search of LLMs by building a training set between pruned model and accuracy, which improves the efficiency of predicting model to be pruned within the search space. In addition, we use the additional information learned by GBDT to prune the search space, and obtain a better performing model under almost the same global pruning ratio. Our approach improves the efficiency and effectiveness of searching within the search space. And with the increase in model evaluation indicators, our method is more transferable and solves the difficulty in adaptability of manually designed pruning features.

\bibliographystyle{ieeetr}
\bibliography{refs}

\end{document}